\documentclass[runningheads]{llncs}
\usepackage{graphicx}

\usepackage{tikz}
\usepackage{comment}
\usepackage{amsmath,amssymb} 
\usepackage{color}
\usepackage{mathrsfs,algorithm,algorithmicx,algpseudocode}
\usepackage{multirow}
\usepackage{multicol}
\usepackage{indentfirst}
\usepackage{subfigure}

\makeatletter
\newcommand{\printfnsymbol}[1]{%
	\textsuperscript{\@fnsymbol{#1}}%
}
\makeatother

\newcommand{\opensupplement}{
	\setcounter{section}{0}
	\renewcommand\thesection{\thechapter.\Alph{section}}
}
\newcommand{\closesupplement}{
	\renewcommand\thesection{\thechapter.\arabic{section}}
}

\begin{document}
	\pagestyle{headings}
	\mainmatter
	\def\ECCVSubNumber{704}  
	
	\title{Search What You Want: Barrier Panelty NAS for Mixed Precision Quantization} 

	\titlerunning{Search What You Want: BP-NAS for Mixed Precision Quantization}
	%
	\author{Haibao Yu\inst{1}\thanks{Indicates equal contributions} \and
		Qi Han\inst{1}\printfnsymbol{1} \and
		Jianbo Li\inst{1,2} \and
		Jianping Shi\inst{1} \and
		Guangliang Cheng\inst{1} \thanks{Indicates equal corresponding authors} \and
		Bin Fan\inst{3} \printfnsymbol{2}}
	\authorrunning{H. Yu et al.}
	%
	%
	\institute{SenseTime Research, Beijing, China
		\and
		Peking University, Beijing, China
		\and
		University of Science and Technology Beijing, Beijing, China\\
		\email{\{yuhaibao, hanqi, shijianping, chengguangliang\}@sensetime.com, jianbo.li@pku.edu.cn, bin.fan@ieee.org}}
	\maketitle
	
	\begin{abstract}
		Emergent hardwares can support mixed precision CNN models inference that assign different bitwidths for different layers. 
		Learning to find an optimal mixed precision model that can preserve accuracy and satisfy the specific constraints on model size and computation is extremely challenge due to the difficult in training a mixed precision model and the huge space of all possible bit quantizations. 
		
		In this paper, we propose a novel soft Barrier Penalty based NAS (BP-NAS) for mixed precision quantization, which ensures all the searched models are inside the valid domain defined by the complexity constraint, thus could return an optimal model under the given constraint by conducting search only one time. The proposed soft Barrier Penalty is differentiable and can impose very large losses to those models outside the valid domain while almost no punishment for models inside the valid domain, thus constraining the search only in the feasible domain. In addition, a differentiable Prob-1 regularizer is proposed to ensure learning with NAS is reasonable. A distribution reshaping training strategy is also used to make training more stable. BP-NAS sets new state of the arts on both classification (Cifar-10, ImageNet) and detection (COCO), surpassing all the efficient mixed precision methods designed manually and automatically. Particularly, BP-NAS achieves higher mAP (up to 2.7\% mAP improvement) together with lower bit computation cost compared with the existing best mixed precision model on COCO detection.
		\keywords{Mixed Precision Quantization, NAS, Optimization Problem with Constraint, Soft Barrier Penalty}
	\end{abstract}

	\section{Introduction}
	
	\par Deep convolutional neural networks (CNNs) have achieved remarkable performance in a wide range of computer vision tasks, such as image classification~\cite{deng2009imagenet,He_2016_CVPR}, semantic segmentation~\cite{zhao2017pyramid,long2015fully} and object detection~\cite{ren2015faster,redmon2016you}. 
	To deploy CNN models into resource-limited edge devices for real-time inference, one classical way is to quantize the floating-point weights and activations with fewer bits, where all the CNN layers share the same quantization bitwidth. However, different layers and structures may hold different sensibility to bitwidth for quantization, thus this kind of methods often causes serious performance drop, especially in low-bit quantization, for instance, 4-bit.
	
	\par Currently, more and more hardware platforms, such as Turning GPUs~\cite{navidia2018gpu} and FPGAs, have supported the mixed precision computation that assigns different quantization bitwidths to different layers. 
	This motivates the research of quantizing different CNN layers with various bitwidths to pursue higher accuracy of quantized models while consuming much fewer bit operations (BOPs).
	Furthermore, many real applications usually set a hard constraint of complexity on mixed precision models that their BOPs can only be less than a preset budget.
	However, it is computational expensive to determine an appropriate mixed precision configuration from all the possible candidates so as to satisfy the BOPs constraint and achieve high accuracy.
	Assuming that the quantized model has $N$ layers and each layer has $M$ candidate options to quantize the weights and activations, 
	the total search space is as large as $(M)^{N}$.
	
	\par To alleviate the search cost, an attractive solution is to adopt weight-sharing approach like DARTS~\cite{Liu2018DARTS} which incorporates the BOPs cost as an additional loss term to search the mixed precision models.
	However, there exist some problems for this kind of methods:
	i) Invalid search. The framework actually seeks a balance between the accuracy and the BOPs cost, may causing the returned mixed precision models outside the BOPs constraint. To satisfy the application requirement, the search process has to be repeated many times by trial-and-error through tuning the BOPs cost loss weight. This is time consuming and often leads to suboptimal quantization.   
	ii) Very similar importance factors. The importance factors corresponding to different bitwidths in each CNN layer often have similar values, which makes it hard to select a proper bitwidth for each layer.
	iii) Unstable training. Empirically, high-bit quantization and low-bit quantization prefer different training strategies. To balance the favors between high-bit and low-bit, the training of mixed precision models maybe fluctuate a lot and cannot achieve high accuracy.
	
	\par To ensure the searched models to satisfy the BOPs constraint, we model the mixed precision quantization task as an constrained optimization problem, and propose a novel loss term called soft barrier penalty to model the complexity constraint. With the differentiable soft barrier penalty, it bounds the search in the valid domain and imposes large punishments to those mixed precision models near the constraint barrier, thus significantly reducing the risk that the returned models do not fulfill the final deploy requirement. 
	Compared to existing methods~\cite{cai2018proxylessnas,wu2019fbnet,wu2018mixed} which incorporate the complexity cost into loss function and tune the corresponding balance weight, our soft barrier penalty method does not need to conduct numerous trial-and-error experiments. 
	Moreover, our search process focuses on the candidates in the valid search space, then the search will be more efficient to approach much more optimal mixed precision model while satisfying the BOPs cost constraint.
	
	\par To make the importance factors more distinguishable, we propose a differentiable Prob-1 regularizer to make the importance factors converge to 0-1 type. With the convergence of the importance factors in each CNN layer, the mixed precision configuration will be gradually determined.
	To deal with the unstable training issues and improve the accuracy of mixed precision model, we use the distribution reshaping method to train a specific float-point model with uniformly-distributed weights and no long-tailed activations, and use the float-point model as the pretrained mixed precision model.
	Experiments show that the pretrained model and uniformization method can make the mixed precision training much more robust and achieve higher accuracy.

	\par In particular, our contributions are summarized as the followings:
	\begin{itemize}
		\item We propose a novel and deterministic method in BP-NAS to incorporate the BOPs constraint into loss function and bound the search process within the constraint, which saves numerous trial and error to search the mixed precision with high accuracy while satisfying the constraint. 
		\item A novel distribution reshaping method is introduced to mixed precision quantization training, which can address the unstable mixed precision training problems while achieve high accuracy.
		\item BP-NAS surpasses the state-of-the-art efficient mixed precision methods designed manually and automatically on three public benchmarks. Specifically, BP-NAS achieves higher accuracy (up to 2.7\% mAP) with lower bit computation compared with state-of-arts methods on COCO detection dataset.
	\end{itemize}
	
	\section{Related Work}
	Mixed precision quantization techniques aims to allocate different weight and activation bitwidth for different CNN layers. Compared to the traditional fixed-point quantization methods, the mixed precision manner can efficiently recover the quantization performance while consuming less bit operations (BOPs). It mainly involves mixed precision quantization training and mixed precision search from the huge search space.
	
	\noindent\textbf{Quantization}
	involves high-bit quantization training and low-bit quantization training. For the former, ELQ~\cite{zhou2017incremental} adopts incremental strategy to fixes part of weights and update the rest in the training, HWGQ~\cite{cai2017deep} introduces clipped ReLU to remove outliers, PACT~\cite{choi2018pact} further proposes adaptatively learns and determines the clipping threshold, Tensorflow Lite~\cite{jacob2017quantization} proposes asymmetric linear scheme to quantize the weights and activations with Integer-arithmetic-only.
	For the later,  Dorefa~\cite{zhou2016dorefa} and TTN~\cite{zhu2016trained} focus on extremely low bit such as one or two bit. Wang et al.~\cite{wang2018two} proposes two-step to train the low-bit quantization. Gu et al.~\cite{gu2019bayesian} proposes Bayesian-based method to optimize the 1-bit CNNs.
	These quantization methods simply assign the same bit for all layers,
	while we consider to apply the uniformization to robustly train the mixed precision quantization model with different bitwidth.
	Shkolnik et al.~\cite{Moran2020robust} also consider the uniformization for robustness but for post-training quantization.
	
	\noindent\textbf{Mixed precision search} focuses on efficiently searching the mixed precision model with high accuracy while consuming less bit operations(BOPs). By treating the bitwidth as operations, several NAS methods could be used for mixed precision search: RL-based method~\cite{wang2018haq} consider to collect the action-reward involving the hard-ware architecture into the loop, and then use these data to train the agent to search better mixed precision model. But it does not consider to use the loss-aware training model data to search the mixed precision.
	Single-Path~\cite{guo2019single} proposes to use evolutionary algorithm to search the mixed precision to achieve high precision and satisfy the constraint.
	But the search will cost huge resources to train numerous of mixed precision models in the early stage.
	
	In this paper, we apply the weight-sharing NAS method into mixed precision search. To the best of our knowledge, our work is mostly related to DNAS~\cite{wu2018mixed}, which also first introduces the weight-sharing NAS into mixed precision search. However there are significant differences: 1) We propose a novel soft barrier penalty strategy to ensure all the search model will be inside the valid domain. 2) We propose an effective differentiable Prob-1 regularizer to make the importance factors more distinguishable. 3) We incorporate distribution reshaping strategy to make the training much more robust. 
	
	\section{Method}
	\par In this paper, we model the mixed precision quantization task as an constrained optimization problem. Our basic solution is to apply DARTS~\cite{Liu2018DARTS} for addressing such optimization problem. A similar solution~\cite{wu2018mixed} for this problem is also to apply DARTS with gumbel softmax trick by incorporating the complexity cost into loss term. However, it actually seeks a balance between the accuracy of the mixed-precision model and the BOPs constraint, making the search process repeate many times by tuning the complexity cost weight until the searched model can satisfy the application requirement. This is time consuming and often leads to suboptimal quantization. To address this issue, we propose a novel loss term to model the BOPs constraint so that it encourages the search conducting in the valid domain and imposes large punishment to those quantizations outside the valid domain, thus significantly reducing the risk that the returned models do not fulfill the final deploy requirement. Our method is termed as BP-NAS since the new loss term is based on the barrier penalty which will be elaborated in Section 3.2. In addition, a Prob-1 regularizer is proposed along with the barrier penalty to facilitate node selection in the supernet constructed by DARTS. Finally, we propose how to train a uniform-like pretrained model to stabilize the training of mixed precision quantization. In the following, we first describe the problem formulation~(Section 3.1) and then detail our solution~(Section 3.2). 
	
	\subsection{Mixed Precision Quantization Search}
	\par Quantization aims to represent the float weights and activations with linearly discrete values. 
	We denote $i$th convolutional layer weights as $\mathbf{W}_i$. 
	For each weight atom $w\in \mathbf{W}_i$, 
	we linearly quantize it into $b_i$-bit as
	\begin{equation}\label{uniform-quantization}
		Q(w;b_i)=[\frac{\text{clamp}(w,\alpha)}{\alpha/({2^{b_i-1}-1)}}]\cdot \alpha/({2^{b_i-1}-1}),
	\end{equation}
	where $\alpha$ is a learnable parameter, and $\text{clamp}(\cdot,\alpha)$ is to truncate the values into $[-\alpha, \alpha]$, $[\cdot]$ is the rounding operation. 
	For activations, we truncates the values into the range $[0,\alpha]$ in a similar way since the activation values are non-negative after the ReLU layer. For simplicity, we use $(m,n)$-bit to denote quantizing a layer with $m$-bit for its weights and $n$-bit for activations.
	
	\par Since different layers in a model may prefer different bitwidths for quantization,
	we consider the problem of mixed precision quantization.
	For a given network $\mathcal{N}=\{L_1, L_2, \cdots, L_{N}\}$, the mixed precision quantization aims at determining the optimal weight bitwidth $\{{b}_1$, ${b}_2$, $\cdots$, ${b}_N\}$ and activation bitwidth $\{{a}_1$, ${a}_2$, $\cdots$, ${a}_N\}$ for all layers from a set of predefined configurations. This is cast into an optimization problem as following,
	\begin{equation}\label{eq: mixed problem}
		\text{MP}^{*} = \mathop{\text{argmin}}_{\text{MP}}\text{Loss}_{val}(\mathcal{N};\text{MP})
	\end{equation}
	where $\text{MP}$ denotes the mixed precision configuration for $\mathcal{N}$, and $\text{Loss}_{val}$ denotes the loss on the validation set.
	
	\par Usually, there are additional requirements on a mixed precision model about its complexity and memory consumption. Bit operations(BOPs\footnote{We also provide an example to demonstrate this BOPs calculation process in supplementary materials. More formulation of BOPs could be seen in  \cite{Chaim2018uniq,Yochai2019towards}, which consider the memory bandwidth and could be more suitable for real application.}) is a popular indicator to formulate these issues. Mathematically, BOPs is formulated as
	\begin{equation}\label{eq:BOPs definetion}
		\mathcal{B}(\mathcal{N};\text{MP})=\sum_{i} \text{FLOP}(L_{i})*{b}_i*{a}_i,
	\end{equation}
	where $\text{FLOP}(L_{i})$ denotes the number of float point operations in layer $L_{i}$, $(b_i,a_i)$-bit denotes weight and activation bitwidths for $L_{i}$ respectively.
	For convenience,  we also refer BOPs to average bit operations which is actually used in this paper,
	\begin{equation}\label{eq:average-bit definetion}
		\mathcal{B}(\mathcal{N};\text{MP})=((\sum_{i} \text{FLOP}(L_{i})*{b}_i*{a}_i)/\sum_{i} \text{FLOP}(L_{i}))^{1/2}.
	\end{equation}
	A typical requirement in mixed precision quantization search is called BOPs constraint to restrict the searched model within a preset budget,
	\begin{equation}\label{eq:BOPs constraint}
		\mathcal{B}(\mathcal{N};\text{MP}) \leq \mathcal{B}_{max}.
	\end{equation}
	
	As a result, the problem of mixed precision quantization search is to find a set of quantization parameters for different layers so as to keep the original model's accuracy as much as possible while fulfilling some complexity constraints, such as the one listed above.
	
	\subsection{BP-NAS for Mixed Precision Quantization Search}
	
	\subsubsection{NAS for Mixed Precision Quantization}
	\par Due to the large cost of training a mixed precision model and the huge mixed precision configurations, it is impractical to train various mixed precision models and select one to satisfy Eq.(\ref{eq: mixed problem}) and Eq.(\ref{eq:BOPs constraint}). Similar to~\cite{wu2018mixed}, we adopt the weight-sharing NAS, i.e., DARTS~\cite{Liu2018DARTS}, to search mixed precision quantization models. 
	\begin{figure}
		\centering
		\includegraphics[width=0.8\textwidth]{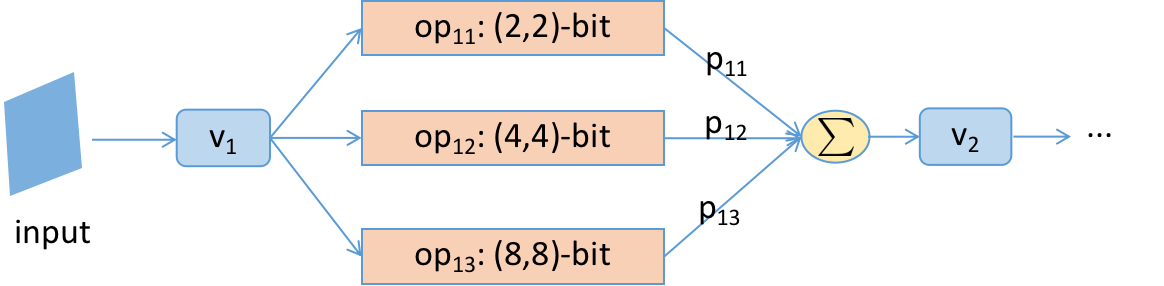}
		\caption{Supernet architecture illustration with (2,2)-bit,(4,4)-bit,(8,8)-bit as candidate bitwidth. $v_i$ represents the input data tensor and each operation share the same input data but different bidwidth.}
		\label{fig:darts_outline}
	\end{figure}
	\par Specifically, we first construct and train a supernet $\mathcal{SN}$, from which the mixed precision model is then sampled.
	The supernet is illustrated in Fig. \ref{fig:darts_outline}.
	In the supernet $\mathcal{SN}$, the edge between node $v_{i}$~(corresponds to the $i$-th layer in the mixed precision model) and $v_{i+1}$ is composed of several candidate bit operations $\{op_{i,j}|_{j=1,\cdots,m}\}$ such as with learnable parameters $\{\theta_{i,j}|_{j=1,\cdots,m}\}$. 
	The output of node $v_{i+1}$ is calculated by assembling all edges as
	\begin{equation}\label{equation:super-net-output}
		v_{i+1} = \sum_{j} p_{i,j}*op_{i,j}(v_{i}),
	\end{equation}
	where $p_{ij}$ denotes the importance factor of the edge $op_{i,j}$ and is calculated with the learnable parameter $\theta_{i,j}$ as
	\begin{equation}\label{equation:probability}
		p_{i,j}=\frac{exp(\theta_{i,j})}{\sum_{l}exp(\theta_{i,l})}.
	\end{equation}
	Once the supernet has been trained, the mixed precision model can be sampled
	\begin{equation}\label{eq:sample}
		\text{MP}^{*}=\{(b_i^{*},a_i^{*})\text{-bit}, i=1,2,\cdots,N\} = \{\text{argmax}_{j}\;p_{i,j} \}
	\end{equation}
	Finally, the sampled mixed precision model is retrained to achieve much better performance. 
	
	\par To take the BOPs cost of the sampled model into consideration, the supernet is trained with the following loss
	\begin{equation}
		Loss = Loss_{val} + \lambda * 	E(\mathcal{SN}).
	\end{equation}
	where BOPs cost of the sampled model is estimated with the supernet as
	\begin{equation}
		\label{eq: expected cost}
		E(\mathcal{SN}) = \sum_{i}\sum_{j}p_{i,j}*\mathcal{B}(op_{i,j}).
	\end{equation}
	Such an estimation is reasonable if the importance factors $p_{i,j}$ for each layer is highly selective, i.e., only one element approaches 1 and others being very small approaching 0. We will introduce later the proposed Prob-1 regularizer to ensure such property of $p_{i,j}$. 
	
	However, optimizing over the above loss only encourages the BOPs being as small as possible while maximizing the accuracy. With a large $\lambda$, it will make the learning focus on the complexity but is very likely to return a less accurate model with much less BOPs than what we want. On the contrary, a small $\lambda$ makes the learning focus on the accuracy and will possibly return a model with high accuracy but with much larger BOPs than what we want. Therefore, it requires numerous trial and error to search the proper mixed precision. This will definitely consume much more time and GPU resources to find a satisfactory mixed precision model. Even though, the obtained model could be suboptimal. 
	
	
	
	Alternatively, to make the returned model satisfying the hard constraint in~(\ref{eq:BOPs constraint}), the mixed precision search is expected to be focused on the feasible space defined by the imposed constraint. In this case, the search is able to return a model fulfilling the constraint and with the better accuracy. Since the search is only conducted in the feasible space, the searching space is also reduced, and so the search is more efficient. We achieve this goal by proposing a novel approach called BP-NAS, which incorporates a barrier penalty in NAS to bound the mixed precision search within the feasible solution space with a differentiable form.

	
	\subsubsection{Barrier Penalty Regularizer}
	The key idea of making the search within the feature space is to develop a regularizer so that it punishes the training with an extremely large loss once the search reaches the bound or is outside the feasible space. For the BOPs constraint in Eq.(\ref{eq:BOPs constraint}), it can be rewritten as a regularization term
	\begin{equation}\label{equation:barrier_losss_1}
		\mathcal{L}_c(\theta)= \begin{cases}
			0 & if \quad E(\mathcal{SN})\le \mathcal{B}_{max} \\
			\infty & otherwise
		\end{cases}
	\end{equation}
	It imposes an $\infty$ loss once the BOPs of the trained supernet falls outside the constraint, while no punishment in the other case. This could be an ideal regularizer as we analyzed before, however, it is non-differentiable and thus can not be used.
	
	\begin{figure}
		\setlength{\abovecaptionskip}{-0.2cm}   %
		\setlength{\belowcaptionskip}{-0.5cm}
		\centering
		\includegraphics[width=0.8\textwidth,height=0.25\textheight]{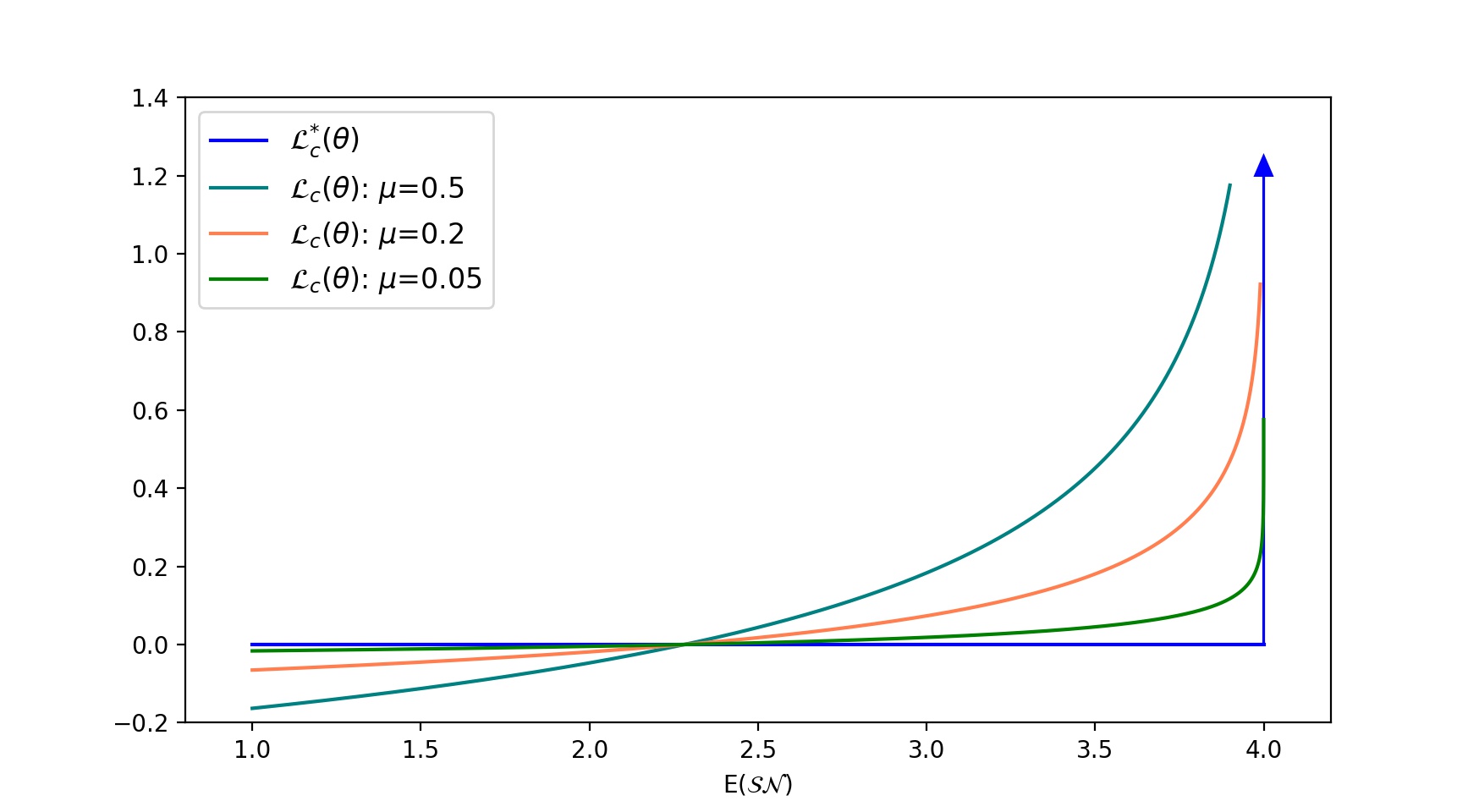}
		\caption{Illustration of the barrier penalty regularizer under $\mathcal{B}_{max}=4$. Blue curves denotes the function $\mathcal{L}_c(\theta)$. Other curves denote $\mathcal{L}^{*}_c(\theta)$ with different $\mu=0.5,0.2,0.05$. With $\mu$ decreasing, the curve of $\mathcal{L}^{*}_c(\theta)$ will be more flatten and is near to 0 in valid interval, and it will be infinity when $E(\mathcal{SN})$ tends to barrier at $E(\mathcal{SN})=\mathcal{B}_{max}$.}
		\label{fig:barrier_function}
	\end{figure}
	Inspired by the innerior method~\cite{alizadeh1995interior} for solving the constrained optimization problem, we approximate $\mathcal{L}_c(\theta)$ by
	\begin{equation}\label{equation:barrier_loss_2}
		\mathcal{L}^{*}_c(\theta)=-\mu log(log(\mathcal{B}_{max} + 1 - E(\mathcal{SN}))),
	\end{equation}
	where $\mu$ should be smaller in the early search stage because there is a large gap between the estimation BOPs of the initial supernet and the BOPs of the final sampled mixed precision model.
	
	Eq.~(\ref{equation:barrier_loss_2}) is differentiable with respect to $E(\mathcal{SN})$ in valid interval $(-\infty, \mathcal{B}_{max})$. As shown in Fig.~\ref{fig:barrier_function}, it approaches the hard constraint loss in Eq.~(\ref{equation:barrier_losss_1}) by decreasing $\mu$ and looks like setting a barrier at $E(\mathcal{SN}) = \mathcal{B}_{max}$. That why we call it the barrier penalty regularizer. For other values of $E(\mathcal{SN})$ in the valid interval, $\mathcal{L}^{*}_c(\theta)$ is near zero, so has very little impact when searching in the feasible solution space.

	\subsubsection{Prob-1 Regularizer}
	\par Since we use the expected BOPs of the supernet to estimate that of the sampled mixed precision model as in Eq.(\ref{eq: expected cost}), this approximation is reasonable only when the supernet gradually converge to the sampled mixed precision model. In other words, the importance factors should gradually be 0-1 type. For this purpose, we introduce the differentiable Prob-1 regularizer.
	
	As the preliminary, we first introduce the Prob-1 function that can be formulated as
	\begin{equation}\label{equation:prob-1 function}
		\begin{aligned}
			& f(\mathbf{x}) = \prod_j(1-x_j), \\ 
			& s.t. \quad \sum x_j=1, 0\leq x_j\leq 1,
		\end{aligned}
	\end{equation}
	where $\mathbf{x}=(x_1,\cdots,x_m)$ is m-dimension vector. For the Prob-1 function, it can be proved that it has the following property,
	\begin{property}\label{theorem:prob-1}
		Prob-1 function $f(x)$ achieves the minimal value if and only if there exists unique $x_j$ to reach 1. 
	\end{property}
	The detailed proof is in the supplementary. Since the importance factors of each layer in the supernet satisfy the constraint of Prob-1 function, we design a Prob-1 regularizer for them with the formula
	\begin{equation}\label{equation:prob-1-regularizer}
		\mathcal{L}_{prob-1}=\sum_{i} \prod_{j}(1-p_{i,j}).
	\end{equation}
	Due to the \textit{Property 1}, this regularizer encourages to learn the 0-1 type importance factors.

	
	\subsubsection{BP-NAS Algorithm}
	The proposed BP-NAS algorithm is summarized in Algorithm~\ref{algorithm:framework}. Basically, it contains three stages: training a supernet, sampling mixed precision from the supernet and retrain the mixed precision model. In the first stage, the network weights $W$ and architecture parameter $\theta$ are trained alternately.
	The network weights in the supernet are updated by the gradient of loss function $\mathcal{L}_1$ on training set,
	\begin{equation}\label{equation:loss-1}
		\mathcal{L}_1 = \mathcal{L}_{train}(W;\theta)
	\end{equation}
	While $\theta$ is updated with the gradient of loss function $\mathcal{L}_2$ on validation set
	\begin{equation}\label{equation:loss-2}
		\mathcal{L}_2 = \mathcal{L}_{val}(W;\theta) + \mathcal{L}^{*}_{c} + \mathcal{L}_{Prob-1}
	\end{equation}
	The differentiation form of loss functions of Eq.(\ref{equation:loss-1})-(\ref{equation:loss-2}) will be described in the supplementary.	After training some epochs, we sample the mixed precision quantization model from the trained supernet according to the importance factors as Eq.(\ref{eq:sample}).
	We retrain the mixed precision models for some epochs to achieve better accuracy.

	Another issue about training mixed precision model is that the training often crashes since training a quantization~(especially low-bit quantization) model is very sensitive and different bit quantization usually favors different training strategies~\cite{jacob2017quantization,choi2018pact}. To alleviate this problem, we use distribution reshaping to train a float-point model with uniformly-distributed weights and no long-tailed activations as the pretrained model, which is helpful for robust training under different bit quantizations. For each weight atom $w$ in the $i$th layer's weight $\mathbf{W}_i$, the distribution reshaping simply uses the clipping method with the formula
	\begin{equation}
		\label{scale-clip}
		\text{clip}(w)=\left\{
		\begin{array}{lr}
			T, & w \ge T \\
			w, & w \in (-T, T) \\
			-T, & w \le -T
		\end{array}
		\right.
	\end{equation}
	where
	\begin{equation}\label{scale-clip-threshold}
		T=k \cdot \text{mean}(|\mathbf{W}_i|).
	\end{equation}
	Here we set $k=2$ to reshape the weight distribution uniform-like.
	We use the Clip-ReLU as in PACT~\cite{choi2018pact} to quantize the activations.
	Similar robust quantization work with distribution reshaping could be seen in~\cite{Yu2020low,Moran2020robust}.
	
	\setlength{\abovecaptionskip}{-0.2cm}   %
	\setlength{\belowcaptionskip}{-0.5cm}
	\begin{algorithm}[ht]
		\caption{Barrier Penalty for Neural Network Search}
		\label{algorithm:framework}
		\begin{algorithmic}[1]
			\Require \textsl{supernet with weight $\textbf{W}$ and architecture parameter $\theta$, BOPs constraint $\mathcal{B}_{max}$, maximal epoch $epoch_{max}$}
			\Ensure \textsl{the architecture with high accuracy under architecture constraints}
			\item $\theta:$ initialize the architecture parameter $\theta$
			\For{epoch = 1:$epoch_{max}$}
			\State update the weight $\textbf{W}$ of the supernet with $\mathcal{L}$ in Eq.(\ref{equation:loss-1})
			\State update the architecture parameter $\theta$ in Eq.(\ref{equation:loss-2}) by computing $\partial (\mathcal{L}_{val}(\textbf{W};\theta) + \mathcal{L}^{*}_{c} + \mathcal{L}_{Prob-1})/\partial \theta$
			\EndFor 
			\State \Return mixed precision model $\text{MP}^{*}$ sampled from the supernet
		\end{algorithmic}
	\end{algorithm}
	
	\section{Experiments}
	We implement the proposed BP-NAS for mixed precision search on image classification~(Cifar-10, ImageNet) and object detection~(COCO).
	The search adopts a block-wise manner, that all layers in the block share the same quantization bitwidth.
	We present all the block-wise configurations for mixed-precision quantization model in the supplementary material.

	\subsection{Cifar-10}\label{section:exp-cifar10}
	We implement the mixed precision search with ResNet20 on Cifar-10, and conduct three sets of experiments under different BOPs constraints $\mathcal{B}_{max}$= 3-bit, 3.5-bit and 4-bit. 
	Compared to 32-bit float point model, the three constraints respectively have $114\times$, $85\times$ and $64\times$ bit operations compression ratio($\mathcal{B}$-Comp).
	
	We construct the supernet whose macro architecture is the same as ResNet20. Each block in the supernet contains \{(2,3), (2,4), (3,3), (3,4), (4,4), (4,6), (6,4), (8,4)\}-bit quantization operations. To train the supernet, we randomly split 60\% of the Cifar-10 training samples as training set and others as validation set, with batch size of 512.
	To train the weight $W$, we use SGD optimizer with an initial learning rate 0.2 (decayed by cosine schedule), momentum 0.9 and weight decay 5e-4. To train the architecture parameter $\theta$, we use Adam optimizer with an initial learning rate 5e-3 and weight decay 1e-3.
	We sample the mixed precision from supernet as Eq. (\ref{eq:sample}),
	and then retrain the sampled mixed precision for 160 epochs and use cutout in data augmentation. Other training settings are the same as the weight training stage.
	To reduce the deviation caused by different quantization settings and training strategies, and to make a fair comparison with other mixed precision methods like HAWQ~\cite{dong2019hawq}, we also train \{(32,32), (2.43{\tiny MP}, 4)\footnote{2.43{\tiny MP} uses the mixed precision quantizations searched by HAWQ~\cite{dong2019hawq}.}, (3,3)\}-bit ResNet20 with our quantization training strategies.
	We remark these results as ``Baseline''.
	
	\begin{table}
		\centering
		\renewcommand\tabcolsep{3pt}
		\caption{Quantization results of ResNet20 on Cifar-10. ``w-bits'' and ``a-bits'' represent the bitwidth for weights and activations.
			``W-Comp'', ``B-Comp'' and ``Ave-bit' denote the model size compression ratio, the bit operations compression ratio, and the average bit operations of mixed precision model, respectively.
			``MP'' denotes mixed precision.}\label{ta:resnet20-cifar10}
		\begin{tabular}{lcccccc}
			\hline
			Quantization & w-bits & a-bits & Top-1 Acc. & W-Comp & B-Comp & Ave-bit \\
			\hline
			\hline
			Baseline & 32 & 32 & 92.61 & $1.00\times$  & $1.00\times$ & 32-bit \\
			\hline
			Baseline & 2.43 \tiny MP & 4 & 92.12 &  $13.11\times$  & $64.49\times$ & 4.0-bit \\
			DNAS~\cite{wu2018mixed} & MP & 32 & 92.00 &  $16.60\times$  & $16.60\times$ & 7.9-bit\\ 
			DNAS~\cite{wu2018mixed} & MP & 32 & 92.72 &  $11.60\times$  & $11.60\times$ & 9.4-bit \\
			HAWQ~\cite{dong2019hawq} & 2.43 \tiny MP & 4 & 92.22 &  $13.11\times$  & $64.49\times$ & 4.0-bit \\
			\textbf{BP-NAS} \tiny \text{$\mathcal{B}_{max}$=4-bit} & 3.14 \tiny MP & MP & \textbf{92.30} & $10.19\times$ & \textbf{81.53}$\times$ &$\textbf{3.5}$-bit \\
			\hline
			\textbf{BP-NAS} \tiny \text{$\mathcal{B}_{max}$=3.5-bit} & 2.86 \tiny MP & MP & \textbf{92.12} & $10.74\times$ & \textbf{95.61}$\times$ & $\textbf{3.3}$-bit\\
			\hline
			Baseline & 3 & 3 & 91.80 & $10.67\times$  & $113.78\times$ & 3.0-bit  \\
			Dorefa~\cite{zhou2016dorefa} & 3 & 3 & 89.90 &  $10.67\times$  & $113.78\times$ & 3.0-bit \\
			PACT~\cite{choi2018pact} & 3 & 3 & 91.10 &  $10.67\times$  & $113.78\times$ & 3.0-bit \\
			LQ-Nets~\cite{zhang2018lq} & 3 & 3 & 91.60 &  $10.67\times$  & $113.78\times$ & 3.0-bit\\
			\textbf{BP-NAS} \tiny \text{$\mathcal{B}_{max}$=3-bit} & 2.65 \tiny MP & MP & \textbf{92.04} & 12.08$\times$ & $\textbf{116.89}\times$ & $\textbf{2.9}$-bit \\
			\hline
		\end{tabular}
	\end{table}
	The results are shown in Table \ref{ta:resnet20-cifar10}. Compared with PACT~\cite{choi2018pact} and HAWQ~\cite{dong2019hawq}, though our Baselines achieves lower accuracy with (3,3)-bit and (2 {\tiny MP},4)-bit on ResNet20, the BP-NAS still surpasses other state-of-the-art methods. Specifically, HAWQ~\cite{dong2019hawq} achieves 92.22\% accuracy with $64.49\times$ compression ratio, whereas BP-NAS achieves 92.30\% accuracy with much higher compression ratio up to $81.53\times$. In addition, our faster version (i.e. BP-NAS with \text{$\mathcal{B}_{max}$=3-bit}) outperforms PACT~\cite{choi2018pact} and LQ-Nets~\cite{zhang2018lq} more than 0.4\%. Note that with different bit computation cost constraints, BP-NAS can search the mixed precision that could not only satisfy the constraint but also achieve even better performance.
	
	\subsection{ImageNet}
	In the ImageNet experiment, we search the mixed precision for the prevalent ResNet50. To speedup the supernet training, we randomly sample 10 categories with 5000 images as training set and with 5000 images as validation set, and set candidate operations as \{(2,4),(3,3),(3,4),(4,3),(4,4),(4,6),(6,4)\}-bit.
	We sample the mixed precision from the supernet and then transfer it to ResNet50 with 1000 categories.
	We retrain the mixed precision for 150 epochs with batch size of 1024 and label smooth on 16 GPUs.
	We also train the (32,32)-bit (3,3)-bit and (4,4)-bit ResNet50 with the same quantization training strategies, and remark the results as ``Baseline''.
	We report the quantization results in Table \ref{ta:resnet50-imagenet}. 
	\begin{table}
		\centering
		\caption{Quantization results of ResNet50 on ImageNet dataset.}
		\label{ta:resnet50-imagenet}
		\begin{tabular}{lccccccc}
			\hline
			Quantization & w-bits & a-bits & Acc.(Top-1) & Acc.(Top-5) & $\mathcal{B}$-Comp & Ave-bit \\
			\hline
			Baseline & 32 & 32 & 77.56\%  & 94.15\% & $1.00\times$ &32-bit \\
			\hline
			\hline
			Baseline & 4 & 4 & 76.02\% & 93.01\% & $64.00\times$ & 4-bit \\
			PACT~\cite{choi2018pact} & 4 & 4 & 76.50\% & 93.20\%  & $64\times$ & 4-bit \\
			HAWQ~\cite{dong2019hawq} & MP & MP & 75.30\% & 92.37\%  & $64.49\times$ & 4-bit \\
			HAQ~\cite{wang2018haq} & MP & MP & 75.48\% & 92.42\% & $78.60\times$ & 3.6-bit \\
			\textbf{BP-NAS} \tiny \text{$\mathcal{B}$=4-bit} & MP & MP & \textbf{76.67}\% & 93.55\%  & \textbf{71.65}$\times$ &$\textbf{3.8}$-bit \\
			\hline		
			Baseline & 3 & 3 & 75.17\% & 92.31\% & $113.78\times$ & 3-bit \\
			Dorefa~\cite{zhou2016dorefa} & 3 & 3 & 69.90\% & 89.20\%  & $113.78\times$ & 3-bit \\
			PACT~\cite{choi2018pact} & 3 & 3 & 75.30\% & 92.60\%   & $113.78\times$ & 3-bit \\
			LQ-Nets~\cite{zhang2018lq} & 3 & 3 & 74.20\% & 91.60\% & $113.78\times$ & 3-bit \\
			\textbf{BP-NAS} \tiny \text{$\mathcal{B}$=3-bit}  & MP  & MP & \textbf{75.71}\% & \textbf{92.83}\%  & $\textbf{118.98}\times$ & $\textbf{2.9}$-bit \\
			\hline
		\end{tabular}
	\end{table}
	\par Under the constraint of \text{$\mathcal{B}_{max}$=3-bit} , the mixed precision ResNet50 searched by our BP-NAS achieves $75.71$\% Top-1 accuracy with compression ratio of $118.98\times$. Though our baseline Top-1 accuracy is inferior to PACT~\cite{choi2018pact}, our mixed precision result surpasses PACT~\cite{choi2018pact} in (3,3)-bit with similar compression ratio. Furthermore, we also present the mixed precision ResNet50 under the constraint \text{$\mathcal{B}_{max}$=4-bit} . Compared to the fixed (4,4)-bit quantization, our mixed precision ResNet50 achieves 76.67\% Top-1 accuracy with similar average bit.
	\par We also compare the quantization results with HAQ~\cite{wang2018haq} and HAWQ~\cite{dong2019hawq}, which also use mixed precision for quantization.
	With similar compression ratio, our mixed precision on ResNet50 under $\mathcal{B}_{max}$=3-bit outperforms HAQ~\cite{wang2018haq} and HAWQ~\cite{dong2019hawq} by 1\% Top-1 accuracy.
	
	\subsection{COCO Detection}
	\par We implement the mixed precision search with ResNet50 Faster R-CNN on COCO detection dataset, which contains 80 object categories and 330K images with 1.5 million object instances.
	Similar to state-of-the-art quantization algorithm FQN~\cite{Li_2019_CVPR}, we also quantize all the convolutional weights and activations including FC layer with fully quantized mode.
	We use the ImageNet to pretrain the backbone and finetune the quantized Faster R-CNN for 27 epochs with batch size of 16 on 8 GPUs. We use SGD optimizer with an initial learning rate 0.1 (decayed by MultiStepLR schedule similar to milestones [16,22,25]), momentum 0.9 and weight decay 1e-4. We report our 4-bit quantization results as well as FQN~\cite{Li_2019_CVPR} in Table~\ref{ta:faster-COCO}. FQN~\cite{Li_2019_CVPR} achieves 0.331 mAP with 4-bit Faster R-CNN and results in 4.6\% mAP loss compared to its float model.
	According to the ``Baseline'', our 4-bit Faster R-CNN achieves 0.343 mAP and outperforms FQN~\cite{Li_2019_CVPR} by 1.2\%.
	Note that our quantization method achieves new state-of-the-art performance on COCO dataset with 4-bit quantization.
	
	\begin{table*}
		\centering
		\renewcommand\tabcolsep{2pt}
		\caption{Quantization results of Faster R-CNN on COCO.}\label{ta:faster-COCO}
		\begin{tabular}{l|c|ccc|cccccc}
			\hline
			\multirow{2}*{Quantization} & \multirow{2}*{input} & \multirow{2}*{w-bits} & \multirow{2}*{a-bits}  & \multirow{2}*{$\mathcal{B}$-Comp} & 
			\multicolumn{6}{c}{mAP} \\
			& & & & & AP & $AP^{0.5}$ & $AP^{0.75}$ & $AP^{S}$ & $AP^{M}$ & $AP^{L}$  \\
			\hline
			Baseline& 800 & 32 & 32 & $1.00\times$ & 0.373 & 0.593 & 0.400 & 0.222 & 0.408 & 0.475 \\
			Baseline& 800 & 4 & 4 & $64.00\times$ & \textbf{0.343} & 0.558 & 0.364 & 0.210 & 0.385 & 0.427 \\
			\hline
			FQN~\cite{Li_2019_CVPR} & 800 & 4 & 4 & $64.00\times$ & 0.331 & 0.540 & 0.355 & 0.182 & 0.362 & 0.436 \\
			\hline
			\textbf{BP-NAS}\tiny$\mathcal{B}_{max}$=4 & 800 & MP & MP & $64.76\times$ & \textbf{0.358} & 0.579 & 0.383 & 0.217 & 0.398 & 0.474 \\
			\hline
		\end{tabular}
	\end{table*}
	As we can see, 4-bit Faster R-CNN still results in significant performance degradation (i.e. 3\% drop). To alleviate this, we implement BP-NAS to search the mixed precision for Faster R-CNN. We utilize the candidate operations from \{(2,4),(3,3),(3,4),(3,5),(4,4),(4,6),(6,4),(4,8)\}.
	Following~\cite{cai2018proxylessnas}, we sample paths by probability to participate the training. 
	To train the supernet, we set the bit operations constraint $\mathcal{B}_{max}$ as 4-bit, which has the same compression ratio as 4-bit quantization.
	We randomly samples 5 categories with 6K images as training set and 4K images as validation set.
	We train the supernet for 40 epochs with batch size of 1.
	Other training settings are the same as the training with ResNet20 on Cifar-10. We train the sampled mixed precision Faster R-CNN with the same training settings as 4-bit quantizaiton. As illustrated in Table \ref{ta:faster-COCO}, the mixed 4-bit Faster R-CNN can achieve 0.358 mAP with bit computation cost compression ratio up to $64.76\times$, which outperforms the original 4-bit Faster R-CNN by 1.5\% mAP. In addition, the mixed 4-bit model only brings in 1.5\% mAP degradation compared to its float-point version, which is a new state-of-the-art on mixed 4-bit community.
	
	\section{Ablation Study}
	We also conduct experiments to illustrate the statistical significance for the mixed precision search, and the effectiveness of Prob-1 regularizer. All experiments are implemented with ResNet20 on Cifar-10 dataset.
	
	\subsection{Efficient Search}\label{sec: efficient search}
	BP-NAS can make the mixed precision search focus on the feasible solution space and return the valid search results with high accuracy.
	We conduct two sets of search experiments for ResNet20 under different BOPs constraint $\mathcal{B}_{max}$=3-bit and $\mathcal{B}_{max}$=4-bit.
	We implement each set of experiments for 20 times respectively, and then we retrain the searched mixed precision models for 160 epoches.
	The accuracy and average bit of the search results are shown in Fig.~\ref{fig:ablation_study_search}.
	
	In Fig.~\ref{fig:ablation_study_search}, the average bit of the search results are all smaller than the preset BOPs constraint. And after retaining for some epoches, many of these mixed precision model can achieve comparable, even higher accuracy than the fixed bit quantization with $\mathcal{B}_{max}$. Take the red circle as an example. They are mixed precision searched with BP-NAS under preset constraint $\mathcal{B}_{max}$=3-bit, and their average bit are all smaller than 3-bit. However, some of them can achieve 91.8\% Top-1 accuracy, which is 3-bit quantization model's accuracy. Even one mixed precision achieves 92.04\% Top-1 accuracy, which surpasses the 3-bit model up to 0.24\% Top-1 accuracy.
	The results show that our BP-NAS is effective for mixed precision quantization with constraint.
	
	Note that if without Barrier Penalty Regularizer, most of the searched bit-widths will tend to be higher bit-width like 8-bit, and the searched mixed-precision models achieve higher Top-1 accuracy up to 92.6\%. Actually, it is natural to allocate higher bit width for each layer without BOPs constraint as higher bit-width leads to higher accuracy.
	
	\subsection{Differentiated Importance Factors}
\begin{figure}[htbp]
		\setlength{\abovecaptionskip}{-0.2cm}   %
		\setlength{\belowcaptionskip}{-0.5cm}
		\centering
		\begin{minipage}[t]{0.5\textwidth}
			\centering
			\includegraphics[width=1\textwidth]{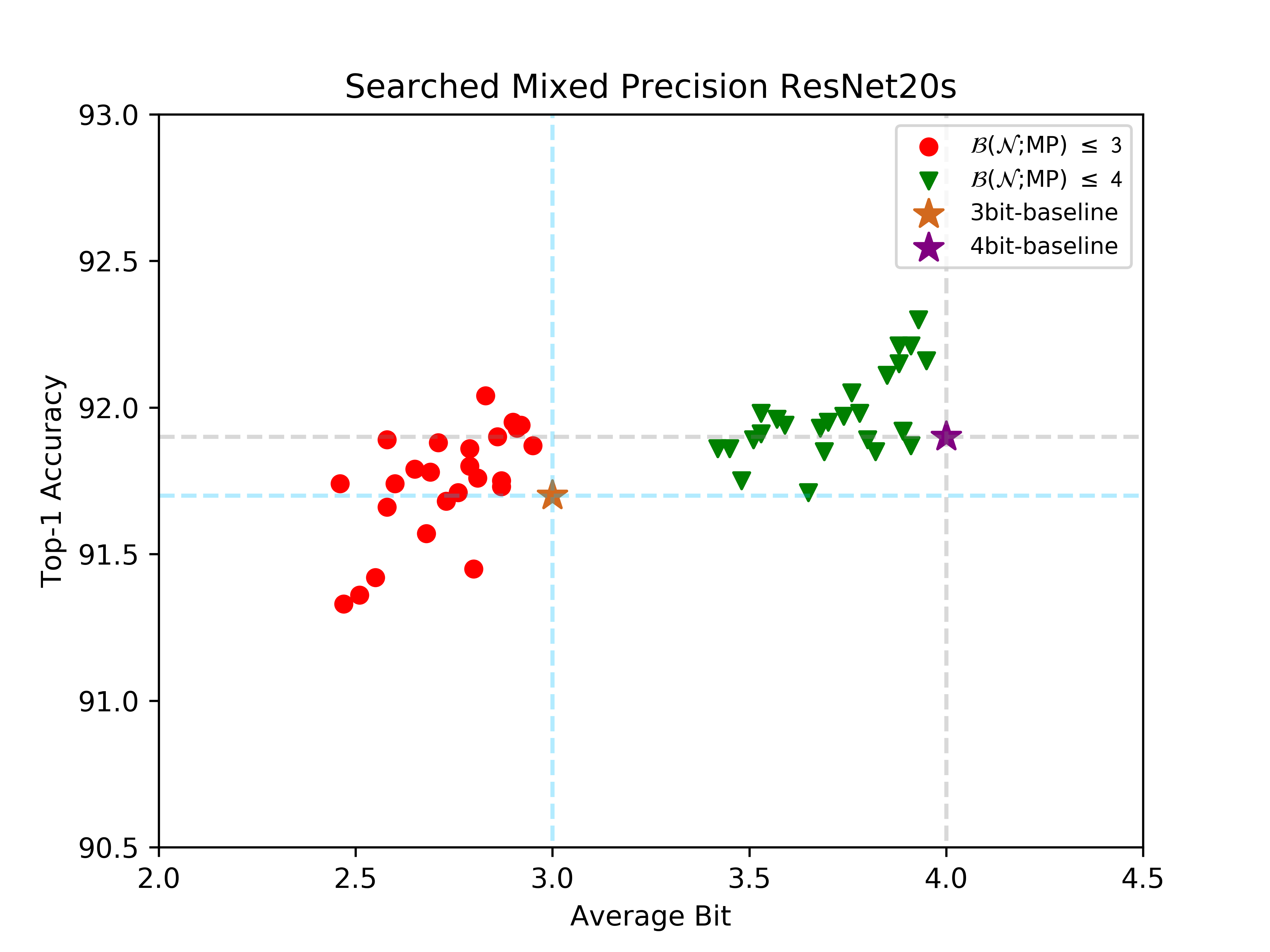}
			\caption{Accuracy and average bit of the search results under different BOPs constraint. Red cicles and green triplet respectively denote the search results under constraints $\mathcal{B}_{max}$=3-bit and $\mathcal{B}_{max}$=4-bit.}\label{fig:ablation_study_search}
		\end{minipage}
		\begin{minipage}[t]{0.45\textwidth}
			\centering
			\includegraphics[width=6cm]{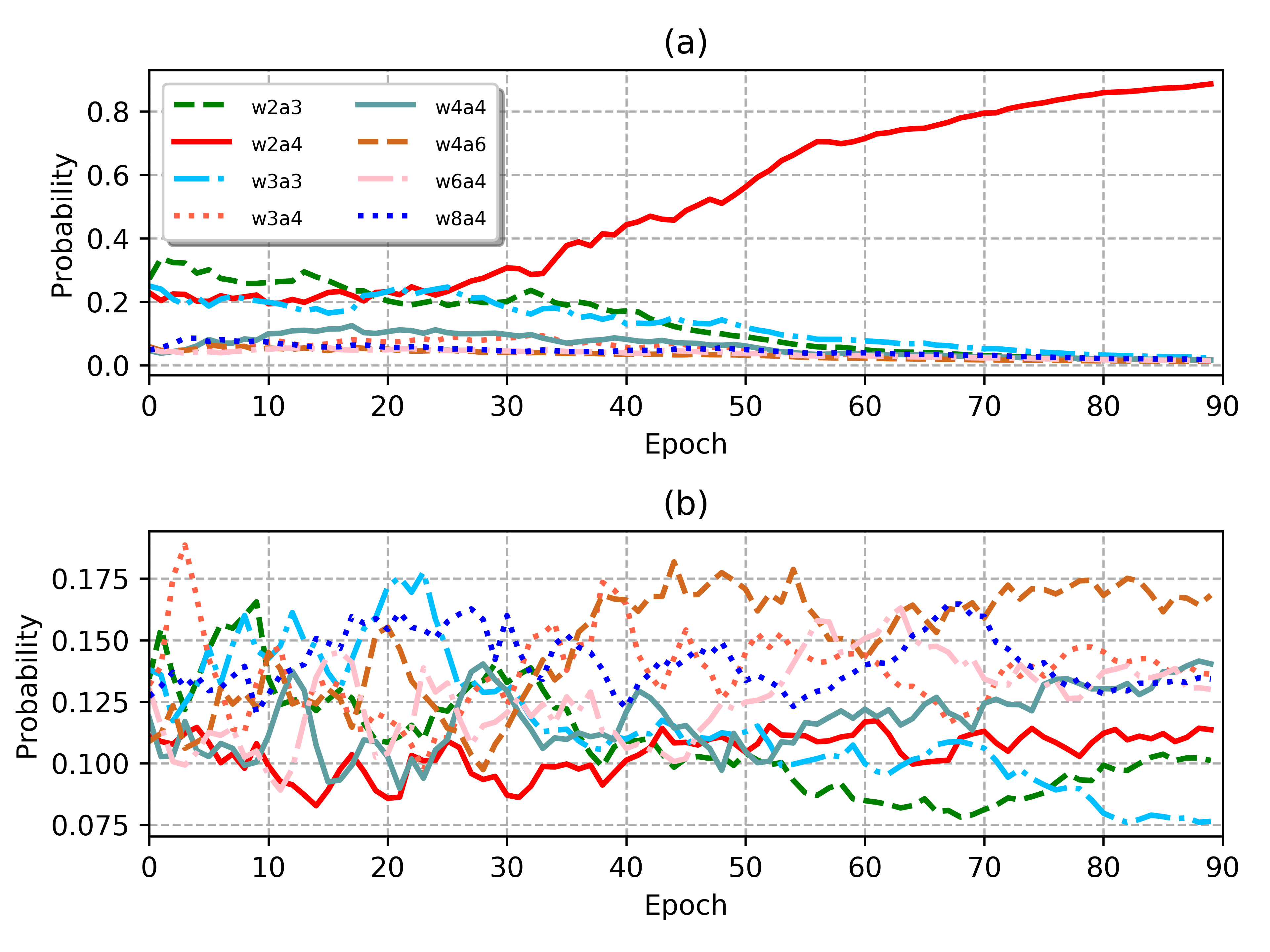}
			\caption{Importance factors evolution. (a) The factors evolution of the first layer with Prob-1 regularizer; (b) The factors evolution of the second layer without the regularizer.}\label{fig:ablation_prob-1}
		\end{minipage}
	\end{figure}

	We conduct two mixed precision search experiments with and without Prob-1 regularizer, to verify whether the Prob-1 regularizer will differentiate the importance factors.
	Both mixed precision search is under the constraint $\mathcal{B}_{max}$=3-bit.
	In Fig.~\ref{fig:ablation_prob-1}, we present the importance factors evolution of the first layer in the two supernets. The red curve in Fig.~\ref{fig:ablation_prob-1}(a) tends to be 1 and other curves tend to 0 after several epoches. While in Fig.~\ref{fig:ablation_prob-1}(b), all the curves have similar value, indicating that these importance factors are similar. 
	Note that our Prob-1 regularizer does not disturb the mixed precision search in the early search stage.
	
	\section{Conclusion}
	In this paper, a novel soft Barrier Penalty Neural Architecture Search (BP-NAS) framework is proposed for mixed precision quantization, which ensures all the search models will satisfy the application meet or predefined constraint. In addition, an effective differentiable Prob-1 regularizer is proposed to differeentiate the importance factors in each layer, and  the distribution reshaping strategy will make the training much more robust and achieve higher performance. Compared with existing gradient-based NAS, BP-NAS can guarantee that the searched mixed precision model within the constraint, while speedup the search to access to the better accuracy. Extensive experiments demonstrate that BP-NAS outperforms existing state-of-the-art mixed-precision algorithms by a large margin on public datasets.
	
	\section*{Acknowledgements}
	We thank Ligeng Zhu for the supportive feedback, and Kuntao Xiao for the meaningful discussion on solving constrained-optimization problem. This work is supported by the National Natural Science Foundation of China~(61876180), the Beijing Natural Science Foundation~(4202073), the Young Elite Scientists Sponsorship Program by CAST~(2018QNRC001).
	
	\clearpage
	%
	%
	\bibliographystyle{splncs04}
	\bibliography{ref}
	
	\clearpage
	\opensupplement
	\textbf{\huge Supplementary Material}
	
	\begin{abstract}
		Section~\ref{section:BOPs-cal} gives an example explaining how BOPs is calculated. Section~\ref{section:property-1} gives the detail proof for the Property 1. In Section~\ref{section: deriv}, it gives the derivative forms of the  Prob-1 regularizer and barrier penalty.
		In Section~\ref{section:architecture}, we provide the searched mixed precision configurations for ResNet20 on Cifar-10 and ResNet50 Faster R-CNN on COCO. Finally, Section~\ref{section: robust} discusses the effectiveness of our distribution reshaping method on mixed precision training.
	\end{abstract}
	
	\section{Example of the BOPs calculation}\label{section:BOPs-cal}
	Let us consider a convolutional layer with $b$-bit weights and $a$-bit activations. Assuming its size is $Ci\times Co\times K\times K$ (where $Ci$=input channel, $K$=kernel size, $Co$=output channel) and the output size is $1\times Co\times H\times W$, then for a single element of the output, it consists of $Ci\times K^{2}$ multiplication operations and 
	$Ci\times K^{2}$ addition operations. Each multiplication operation involves $b\times a$ bit operations, and each addition operation involves $b + a + log_2(Ci\times K^{2})$ bit operations. To have a fair comparison with SOTA work like [3,4], we aslo keep the multiplication and addition sharing the same number of bit operations as $b\times a$.
	Then this convolutional layer yields a total number of bit operations $BOPs\approx FLOPs\times b
	\times a$.
	Moreover, to facilitate our comparison with fixed-precision, we refer $BOPs$ to the average number of bit operations as $(BOPs/FLOPs)^{1/2}$. With a preset bit budget $B_{max}$, the real BOPs budget is $B_{max}^{2}\times FLOPs$.
	
	\section{Proof of Property 1}\label{section:property-1}
	\begin{property}\label{theorem:prob-1}
		Prob-1 function $f(\mathbf{x})$ achieves the minimal value if and only if there exists unique $x_j$ to reach 1,
		where $\mathbf{x}=(x_1, \cdots, x_m)$ is m-dimension vector and
		\begin{equation}\label{equation:prob-1 function}
			\begin{aligned}
				& f(\mathbf{x}) = \prod_j(1-x_j), \quad s.t.~\sum x_j=1, 0\leq x_j\leq 1.
			\end{aligned}
		\end{equation}
	\end{property}
	\begin{proof}
		Firstly, we prove that if there exists unique $x_j$ to reach 1 then $f(\mathbf{x})$ achieves the minimal value. 
		For any $x_j$, it satisfies that $0\leq x_j\leq 1$, such that $0\leq(1-x_j)\leq1$.
		Thus $f(\mathbf{x})$ is always larger than or equal to $0$. 
		When there exists $x_j$ to reach 1, $f(\mathbf{x})$ reaches 0, that means that $f(\mathbf{x})$ achieves the minimal value 0.
		
		Secondly, we prove that if $f(\mathbf{x})$ achieves the minimal value then there exists unique $x_j$ to reach 1.
		Clearly when $f(\mathbf{x})$ achieves the minimal value 0, there exists $1-x_j$ to reach 0.
		Since $0\leq x_j\leq 1$ and $\sum x_j=1$, all other $x_{i,i\neq j}$ is equal to 0.
		Thus there exists unique $x_j$ to reach 1.
	\end{proof}
	
	We also introduce a simple example to illustrate the Property \ref{theorem:prob-1} as follows
	\begin{equation}
		\begin{aligned}
			& f(\mathbf{x}) = (1-x_1)(1-x_2), \\
			& s.t. \quad x_1+x_2=1, 0\leq x_1,x_2\leq 1,
		\end{aligned}
	\end{equation}
	where $\mathbf{x}=(x_1,x_2)$. By replacing $x_2$ as $1-x_1$ we can rewrite the problem as
	$$f(\mathbf{x}) = (1-x_1)x_1, \quad s.t. \quad 0\leq x_1\leq 1.$$ Obviously, $f(\mathbf{x})$ achieves minimum if and only if $x_1$ is assigned to be 0 or 1.
	
	\section{Derivatives of Prob-1 Regularizer and Barrier Penalty}\label{section: deriv}
	\subsection{Derivative of Barrier Penalty}
	Consider the barrier penalty 
	\begin{equation}\label{equation:barrier_loss_2}
		\mathcal{L}^{*}_c(\theta)=-\mu log(log( \mathcal{B}_{max}+1-E(\mathcal{SN}))),
	\end{equation}
	where $E$ denotes the expected complexity cost $\mathcal{F}(\mathcal{SN};\theta)$ of the supernet. 
	
	We first take the derivative of the barrier penalty about $E(\mathcal{SN})$ as $\frac{\partial \mathcal{L}^{*}_c}{\partial E}$.
	\begin{equation}
		\begin{aligned}
			\frac{\partial \mathcal{L}^{*}_c}{\partial E}
			& =\frac{\partial (-\mu log(log( \mathcal{B}_{max}+1-E(\mathcal{SN}))))}{\partial E} =-\mu\frac{\partial log(log( \mathcal{B}_{max}+1-E(\mathcal{SN})))}{\partial E} \\
			& = \frac{-\mu}{log( \mathcal{B}_{max}+1-E(\mathcal{SN}))} \frac{\partial log( \mathcal{B}_{max}+1-E(\mathcal{SN}))}{\partial E} \\
			& = \frac{-\mu}{log( \mathcal{B}_{max}+1-E(\mathcal{SN}))} \frac{1}{ \mathcal{B}_{max}+1-E(\mathcal{SN})}\frac{\partial ( \mathcal{B}_{max}+1-E(\mathcal{SN}))}{\partial E} \\
			& = \frac{\mu}{log( \mathcal{B}_{max}+1-E(\mathcal{SN}))( \mathcal{B}_{max}+1-E(\mathcal{SN}))} 
		\end{aligned}
	\end{equation}
	
	Then, according to the chain rule, we have
	\begin{equation}
		\frac{\partial \mathcal{L}^{*}_c}{\partial \theta} = \frac{\partial \mathcal{L}^{*}_c}{\partial E} \frac{\partial E}{\partial \theta}=\frac{\mu}{log( \mathcal{B}_{max}+1-E(\mathcal{SN}))( \mathcal{B}_{max}+1-E(\mathcal{SN}))} \frac{\partial E}{\partial \theta}.
	\end{equation}
	
	Here $\frac{\partial E}{\partial \theta}$ has been discussed in~
	\cite{cai2018proxylessnas,Liu2018DARTS}.
	
	\subsection{Derivative of Prob-1 Regularizer}
	For the $i$th block of the supernet, $p_{i,j}$ represents the importance of the $j$th operation.
	Since $\{p_{i,j}\}$ is obtained by Softmax,
	we have $\sum_j p_{i,j}=1$ and $0\leq p_{i,j}\leq 1$.
	
	For the Prob-1 regularizer with the following form,
	\begin{equation}
		\mathcal{L}_{prob-1}=\sum_{l} \prod_{m}(1-p_{l,m}).
	\end{equation}
	
	The partial derivative $\frac{\partial \mathcal{L}_{prob-1}}{\partial p_{i,j}}$ is
	\begin{equation}
		\begin{aligned}
			\frac{\partial \mathcal{L}_{prob-1}}{\partial p_{i,j}} 
			& = \frac{\partial (\sum _l  \prod_{m}(1-p_{l,m})}{\partial p_{i,j}}
			=\frac{\partial (\sum _{l\neq i}  \prod_{m}(1-p_{l,m})+ \prod_{m}(1-p_{i,m}))}{\partial p_{i,j}} \\
			& =\frac{\partial ( \prod_{m}(1-p_{i,m}))}{\partial p_{i,j}}
			=\frac{\partial ( \prod_{m\neq j}(1-p_{i,m})*(1-p_{i,j}))}{\partial p_{i,j}}\\
			& = \prod_{m\neq j}(1-p_{i,m})*(-1)
		\end{aligned}
	\end{equation}
	
	\section{Mixed Precision Configuration}\label{section:architecture}
	In this section, we provide the exact mixed precision configurations for different blocks for ResNet20 on Cifar-10 as well as ResNet50 Faster R-CNN on COCO. 
	\begin{table}
		\centering
		\renewcommand\tabcolsep{3pt}
		\caption{Mixed precision configuration for ResNet20 on Cifar-10. We abbreviate block type as “B-Type”. “Params” represents the number of weight. “w-bits” and “a-bits” represent the bitwidths for weights and activations, respectively. We report three mixed precision configurations under different BOPs constraints $\mathcal{B}_{max}$ = 4-bit, 3.5-bit, 3-bit. 
			We quantize the first convolutional layer into 8-bit.}
		\begin{tabular}{lccccccccc}
			\hline
			\multirow{2}*{Block} & \multirow{2}*{B-Type} & \multirow{2}*{Params} & \multirow{2}*{FLOPs}  & \multicolumn{2}{c}{$\mathcal{B}_{max}$=3} & \multicolumn{2}{c}{$\mathcal{B}_{max}$=3.5} & \multicolumn{2}{c}{$\mathcal{B}_{max}$=4} \\
			& & & & w-bit & a-bit & w-bit & a-bit & w-bit & a-bit \\
			\hline
			\hline
			Block 0 & Conv & 4.32$\times10^{2}$ & 4.42$\times10^{6}$ & 8 & 8 & 8 & 8 & 8 & 8 \\
			Block 1  & BasicBlock & 4.61$\times10^{3}$ & 4.72$\times10^{6}$ & 3 &  3 &  3 &  3 & 6 & 4\\
			Block 3  & BasicBlock & 4.61$\times10^{3}$ & 4.72$\times10^{6}$ & 3 &  3 &  2 &  4 & 4 & 4\\
			Block 3  & BasicBlock & 4.61$\times10^{3}$ & 4.72$\times10^{6}$ & 3 &  3 &  4 &  4 & 4 & 4 \\
			Block 4  & BasicBlock & 1.38$\times10^{4}$ & 3.54$\times10^{6}$ & 3 &  3 &  4 &  4 & 4 & 3\\
			Block 5  & BasicBlock & 1.84$\times10^{4}$ & 4.72$\times10^{6}$ & 2 &  4 &  3 &  3 & 3 & 3\\
			Block 6  & BasicBlock & 1.84$\times10^{4}$ & 3.54$\times10^{6}$ & 2 &  4 &  2 &  4 & 2 & 4\\
			Block 7  & BasicBlock & 5.53$\times10^{4}$ & 4.72$\times10^{6}$ & 3 &  3 &  3 &  3 & 3 & 3\\
			Block 8 & BasicBlock & 7.37$\times10^{4}$ & 4.72$\times10^{6}$ & 3 &  3 &  3 &  3 & 3 & 3\\
			Block 9  & BasicBlock & 7.37$\times10^{4}$ & 4.72$\times10^{6}$ & 3 &  3 &  3 & 3 & 3 & 3\\
			\hline
		\end{tabular}
	\end{table}
	
	\begin{table}
		\centering
		\renewcommand\tabcolsep{3pt}
		\caption{Mixed precision configuration for ResNet50 Faster R-CNN on COCO.
			We abbreviate block type as "B-Type", number of blocks as "N-Block". All layers in the same block share the same quantization bitwidth.}
		\begin{tabular}{lcccc}
			\hline
			Part & B-Type & N-Block & w\_bit & a\_bit \\
			\hline
			backbone.layer0 & Conv & 1 & 4 & 8 \\
			backbone.layer1 & Bottleneck & 3 & [4,3,3] & [4,5,5] \\
			backbone.layer2 & Bottleneck & 4 & [3,4,4,4] & [5,4,3,4] \\
			backbone.layer3 & Bottleneck & 6 & [4,4,4,3,4,4] & [6,3,4,5,4,3] \\
			backbone.layer4 & Bottleneck & 3 & [3,4,4] & [5,4,6] \\
			neck & Conv & 8 & [4,4,4,4,4,4,4,4] & [4,4,4,4,4,4,4,4] \\
			roi\_head & Conv & 3 & [4,4,4] & [4,4,4] \\
			bbox\_head & Linear & 4 & [4,4,4,4] & [4,4,4,4] \\
			\hline
		\end{tabular}
	\end{table}
	
	\section{Robust Training}\label{section: robust}
	In this section, we discuss the distribution reshaping strategy, which facilitates mixed precision training more robust and achieves higher accuracy.
	
\begin{figure}
		\centering
		\includegraphics[width=1.0\textwidth, height=0.3\textheight]{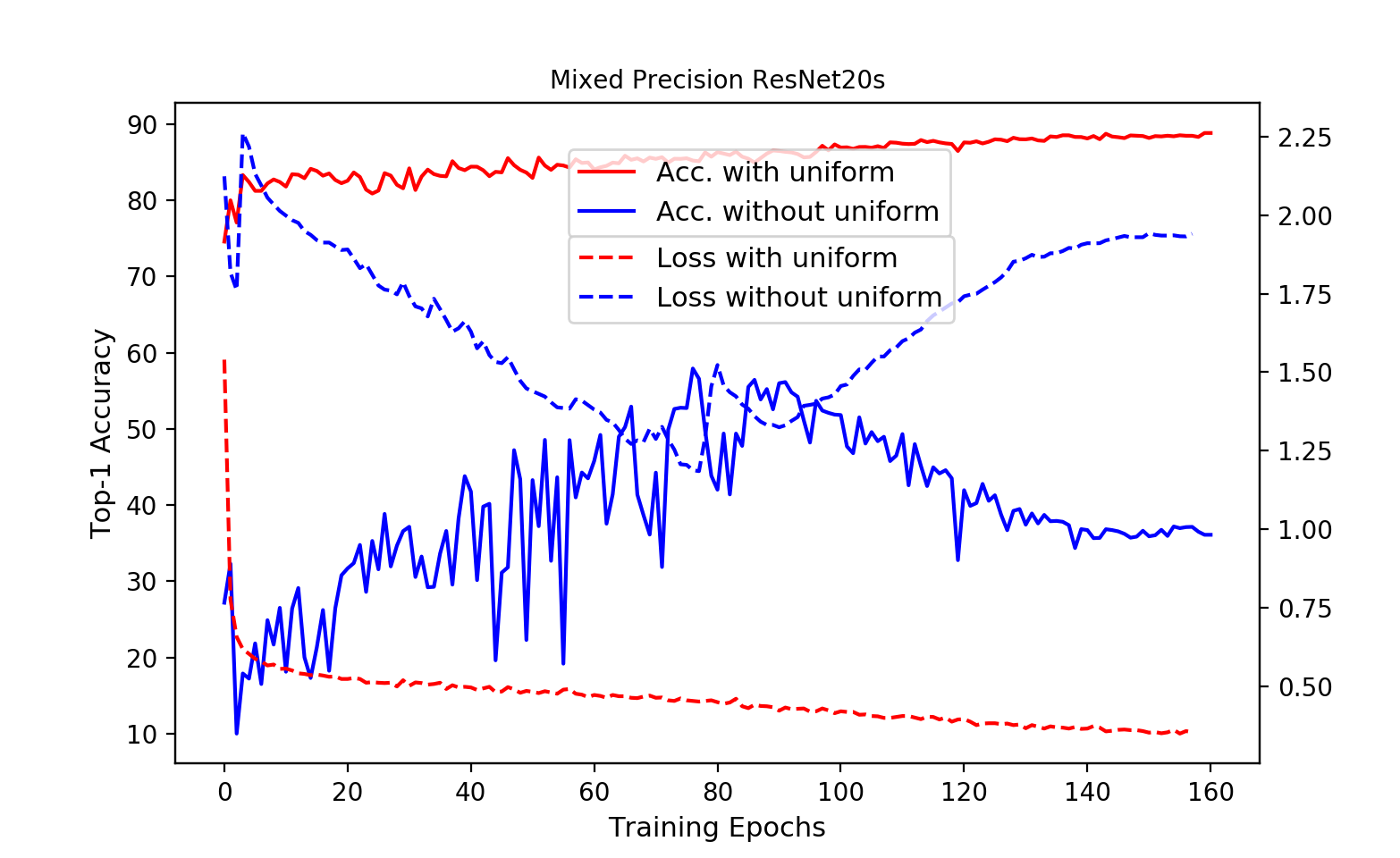}
		\caption{Training loss and validation accuracy with or without uniformization.}
		\label{fig:ablation_robust}
	\end{figure}

	We randomly select bitwidth from $\{(1,1),$$ (2,2),$ $(4,4), $$(8,8)\}$ for each block in ResNet20, and construct a mixed precision ResNet20. We train two mixed precision ResNet20 models with the same mixed precision configuration and training strategies, except whether applying uniformization or not. Both of them are trained for 160 epoches.
	Fig.~\ref{fig:ablation_robust} shows the training loss and validation accuracy of the two mixed precision ResNet20 models.
	In Fig. \ref{fig:ablation_robust}, training with uniformization achieves 89\% Top-1 accuracy, while training without uniformization only achieves no more than 60\% Top-1 accuracy, even drops a lot at last. The Fig.~\ref{fig:ablation_robust} clearly demonstrate the robustness in training with uniformization.
	
\begin{figure}[H]
		\centering
		\includegraphics[width=1.0\textwidth, height=0.3\textheight]{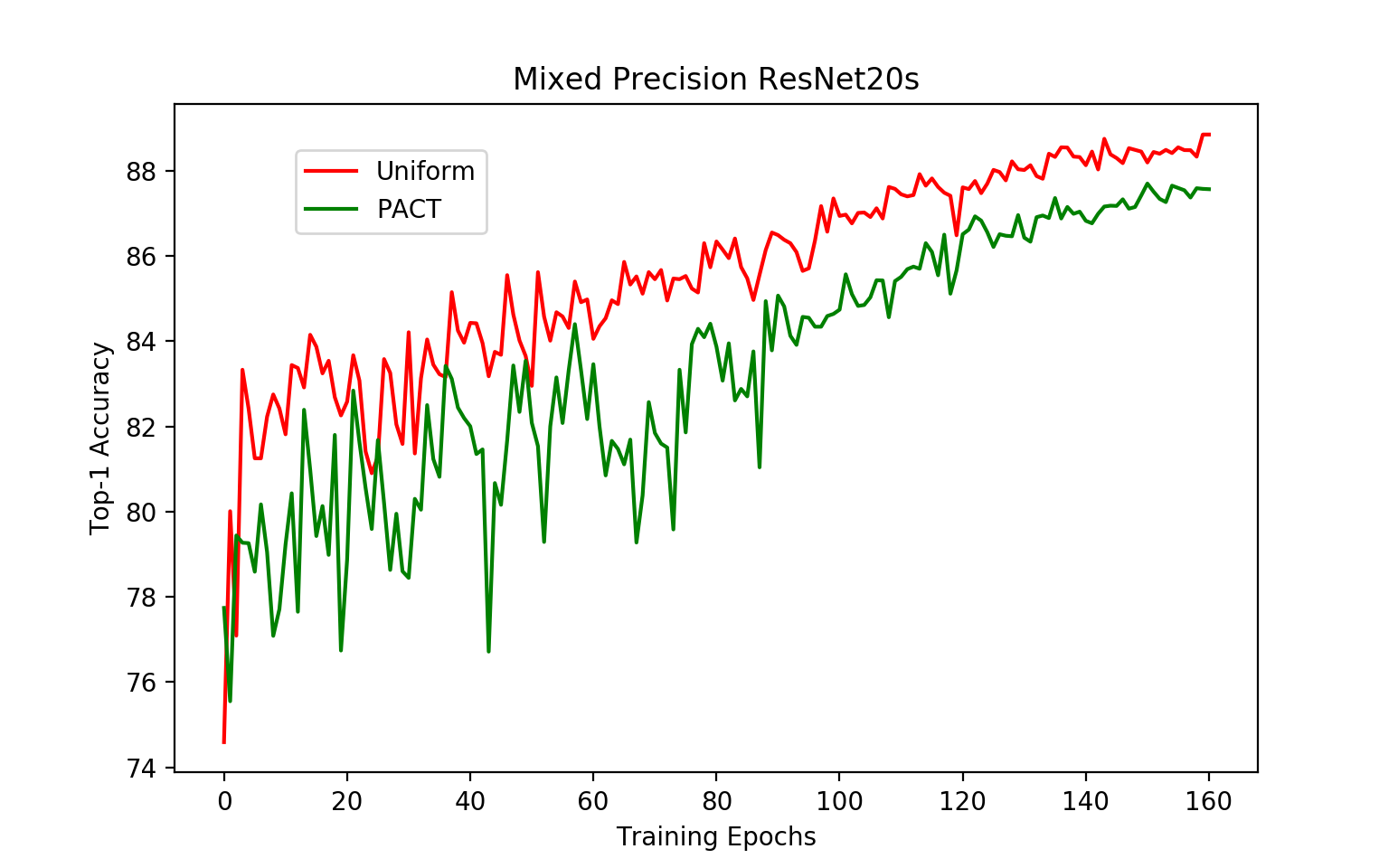}
		\caption{The validation accuracy of mixed precision training with distribution reshaping strategy and PACT~\cite{choi2018pact}.}
		\label{fig:ablation_robust_compare}
	\end{figure}

	Then we compare our distribution reshaping strategy with PACT~\cite{choi2018pact} in Fig. \ref{fig:ablation_robust_compare}.
	Our validation accuracy behaves more robust, and achieves higher Top-1 Accuracy than PACT~\cite{choi2018pact}.
	
	\closesupplement
	
\end{document}